\documentclass[runningheads]{llncs}

\usepackage{eccv}

\usepackage{eccvabbrv}

\usepackage{graphicx}
\usepackage{booktabs}

\usepackage[accsupp]{axessibility}  %

\usepackage[hidelinks]{hyperref}

\usepackage{orcidlink}

\usepackage{url}            %
\usepackage{amsfonts}       %
\usepackage{nicefrac}       %
\usepackage{microtype}      %
\usepackage{xcolor}         %
\usepackage{multirow}  %
\usepackage{array}     %
\usepackage{amsmath}
\usepackage{mathtools}
\usepackage{enumitem}
\usepackage{float}    %
\usepackage{tikz}
\usepackage{subcaption}

\usepackage{xcolor}        %
\usetikzlibrary{calc,matrix}

\definecolor{Gcol8}{RGB}{198,234,191}
\definecolor{Gcol7}{RGB}{122,221,183}
\definecolor{Gcol6}{RGB}{79,189,195}
\definecolor{Gcol5}{RGB}{63,147,201}
\definecolor{Gcol4}{RGB}{74,110,196}
\definecolor{Gcol3}{RGB}{77,80,164}
\definecolor{Gcol2}{RGB}{60,55,116}
\definecolor{Gcol1}{RGB}{45,35,76}

\tikzset{legendbox/.style={
    rectangle, minimum width=0.8em, minimum height=0.8em, draw=none
}}

\usepackage{algorithm}
\usepackage{algpseudocode}
\usepackage{amssymb}
\usepackage{makecell}

\usepackage{lipsum}

\newcommand\blfootnote[1]{%
  \begingroup
  \renewcommand\thefootnote{}\footnote{#1}%
  \addtocounter{footnote}{-1}%
  \endgroup
}

\usepackage{caption}
\usepackage{url}

\newcommand{\revisedROne}[1]{#1}
\newcommand{\revisedRTwo}[1]{#1}
\newcommand{\revisedRThree}[1]{#1}
\newcommand{\revisedRFour}[1]{#1}

\newcommand{\dsk}{d_{\mathrm{sk}}}

\begin{document}

\title{Graph Coloring for Multi-Task Learning} 

\titlerunning{Graph Coloring for Multi-Task Learning}

\author{Santosh Patapati\inst{1}\thanks{Corresponding author.} \and
Ian Noronha\inst{2}}

\authorrunning{S.~Patapati and I.~Noronha}

\institute{Stony Brook University AI Innovation Institute, Stony Brook, NY 11794, USA\\
\email{santosh.patapati@stonybrook.edu}
\and
Purdue University, West Lafayette, IN 47907, USA\\
\email{inoronha@purdue.edu}}

\maketitle

\begin{abstract}
When different objectives conflict with each other in multi-task learning, gradients begin to interfere and slow convergence, 
thereby 
potentially 
reducing the final model's performance. %
To address this, we introduce SON-GOKU, a scheduler that computes gradient interference, constructs an interference graph, and then applies greedy graph-coloring to partition tasks into groups that align well with each other.
At each training step, only one group (color class) of tasks are activated, and the grouping partition is constantly recomputed 
as task relationships evolve throughout training.
By ensuring that each mini-batch contains only tasks that pull the model in the same direction, our method improves the effectiveness of any underlying multi-task learning optimizer without additional tuning.
Since tasks within these groups will update in compatible directions, 
multi-task learning will improve model performance rather than impede it.
Empirical results on six different datasets show that this interference-aware graph-coloring approach consistently outperforms baselines and state-of-the-art multi-task optimizers.
We provide extensive theory %
showing why grouping and sequential updates improve multi-task learning, with guarantees on descent, convergence, and 
the ability to accurately identify what tasks conflict or align.\blfootnote{\revisedRFour{Implementation:} \url{https://anonymous.4open.science/r/SON-GOKU-Impl/}}%

\end{abstract}

\section{Introduction}
Multi-task learning (MTL) trains a single model to solve several tasks simultaneously, sharing knowledge across them to learn more effectively \cite{Caruana1997, Baxter2000}. This allows models to generalize better and converge faster. However, a key issue 
known as negative transfer
arises when tasks don't align very well with each other \cite{Sener2018, Shi2023Recon}. When two tasks push the shared network in different directions their gradients clash, slowing or even reversing learning.
Prior work addresses this issue primarily via (1) gradient manipulation, which reshapes task gradients to reduce conflicts, and (2) loss reweighting, which rescales task objectives to balance their influence. While effective in some specific settings, these strategies typically treat conflict locally at the level of shared-parameter updates and often overlook the evolving global structure of interactions among tasks throughout training. 

Some recent works focus on partitioning tasks into subsets (groups) and updating those groups separately. These approaches have been found to improve accuracy and training stability by forming groups with high measured affinity and then updating one group at a time \cite{Fifty2021, jeong2025selective}. %
Grouping can outperform gradient manipulation and loss reweighting when tasks form clusters with aligned gradients, because each update then reduces direct clashes in the shared layers, lowers gradient variance within the step, and lets compatible tasks reinforce one another while conflicting tasks wait for their turn.

However, grouping methods often face a few key limitations: \revisedROne{(1)} many rely on dense pairwise affinities that grow noisy and costly as the number of tasks rises \cite{Fifty2021,Standley2020TaskGrouping,Sherif_2024}\revisedROne{; (2)} others predetermine or rarely update groups, so they drift as task relations change \cite{wang2024towards, Ruder2017Overview}\revisedROne{; and (3)} several use local heuristics that fail to enforce global compatibility or to specify how groups should rotate over time \cite{zhang2018overview, malhotra2022dropped}.

We present \textbf{SON-GOKU} (\textbf{S}cheduling via \textbf{O}ptimal I\textbf{N}terference-aware \textbf{G}raph-C\textbf{O}loring for Tas\textbf{K} Grouping in M\textbf{U}ltitask Learning).
We measure gradient interference, %
build a graph of tasks from those measurements, greedily color the graph to form non-conflicting compatible task groups, and update one color group per step during training. 
This design addresses the earlier issues. We estimate the interference graph from lightweight minibatch statistics and keep it sparse, which avoids noisy dense matrices and scales to many tasks. We recolor the graph at regular intervals so the groups track changing relations during training. Greedy graph coloring ensures we update only compatible tasks in each step, and the color order gives a simple way to cycle through the groups. Our proposed scheduler does not have to work in isolation and can function 
on top of 
existing loss-reweighting and gradient-manipulation MTL approaches.

In our theoretical analysis (Section \ref{sec:theoretical_analysis}) we show that, 
under standard conditions, SON-GOKU tends to group tasks whose gradients are, on average, aligned within each group, with high probability. We further show that, over a refresh window, sequentially updating these low-conflict groups yields \textit{at least} as much expected descent as a single mixed update, and \textit{strictly more} when between-group interference is sufficiently negative. We also prove that SON-GOKU preserves descent and reaches the usual non-convex SGD rate under 
mild assumptions, with only a small factor that depends on the within-group conflict level. %
In Supp. Materials Section D we discuss the scheduler's amortized time complexity and the tradeoffs it offers between speed and performance. We discuss ways in which practitioners can reduce its time complexity under certain conditions.

Empirical results from experiments demonstrate that SON-GOKU consistently improves outcomes compared to other MTL approaches, especially when SON-GOKU is coupled with existing approaches. Our contributions are as follows:
\begin{itemize}
\item We propose SON-GOKU, an interference-aware scheduler that measures cross-task gradient conflict, builds a conflict graph, colors it to form compatible groups, and activates one group per step. It can be used on top of standard MTL optimizers.
\item We provide theoretical analysis that offers guarantees on SON-GOKU's grouping, convergence, scheduling behavior, and more.
\item Across six datasets, SON-GOKU improves over strong baselines and pairs well with methods like PCGrad, AdaTask, and GradNorm, delivering consistent gains \cite{Yu2020PCGrad, yang2023adatask, Chen2017GradNorm}.
\item We perform an ablation study showing that dynamic recoloring and history-averaged conflict estimates are key contributors to performance. 
\end{itemize}

\section{Related Work}\label{sec:related-work}
\revisedROne{Prior work has identified the phenomenon of gradient interference in multi-task learning and explored several strategies to mitigate it. We group these strategies into four families: (1) \emph{Tuned Loss Weighting}, (2) \emph{Adaptive Loss Weighting}, (3) \emph{Gradient-Level Conflict Mitigation}, and (4) \emph{Empirical Task Grouping}. SON-GOKU falls into family (4).}

Many MTL methods (especially earlier ones) adjust task influence by learning or adapting loss weights. Examples include uncertainty-based scaling \cite{kendall2018multi}, rate-based schemes such as DWA \cite{Liu2019EndToEnd}, and fast bilevel formulations like FAMO \cite{Liu2023FAMO}. 
FAMO in particular is notable for its $\mathcal{O}(1)$ per-step time complexity. These approaches keep all tasks active each step while modulating relative magnitudes. 
A completely different approach, which emerged in 2018 with MGDA \cite{Sener2018}, focuses on updating shared-parameter \emph{update directions} to mitigate interference \cite{lin2019pareto}. Methods like PCGrad \cite{Yu2020PCGrad}, CAGrad \cite{Liu2021CAGrad}, and MGDA \cite{Sener2018} modify the geometry of the shared update 
to reduce cross-task conflicts while still updating all tasks each step.
A smaller body of work forms subsets of tasks to update together, using offline affinity estimation or training-dynamics signals \cite{Fifty2021,Standley2020TaskGrouping,wang2024towards,Sherif_2024}. \revisedRTwo{See Supp. Materials Section Q for additional analysis of non-conflict task grouping.} Most recently, Selective Task Group Updates proposes online grouping with sequential updates, 
reporting that update order can influence task-specific learning \cite{jeong2025selective}. %
We also distinguish SON-GOKU from GO4Align and EXTRA. GO4Align studies task imbalance and alignment, and EXTRA studies task tradeoffs, while SON-GOKU schedules tasks using measured gradient conflict. In contrast to these methods, SON-GOKU changes which tasks are updated together at each step rather than only changing task weights or analyzing tradeoffs \cite{Shen2024GO4Align,zhou2026exploring}.
\revisedROne{SON-GOKU differs in mechanism from existing approaches (Section \ref{sec:proposedapproach}). It complements loss reweighting and gradient surgery, and we provide explicit guarantees on descent, convergence, and graph partition recovery.}
An expanded discussion \revisedROne{and commentary of related work} is provided in Supp. Materials Section M.

\section{Problem Setup}
We formalize multi-task learning (MTL) \cite{Caruana1997} as optimizing a shared network while activating only a subset of tasks at each step. Each task contributes a loss whose gradients may align or conflict. We quantify conflict using (the negative of) cosine similarity, embed tasks in a conflict graph, and later use that graph to derive a schedule (see Supp. Materials Section P for a unique, modular, formulation and results with alternative measures of affinity). This section fixes notation and states the optimization goal that the proposed approach addresses.

\subsection{Data and Notation}
Let $\mathcal T=\{T_1,\dots ,T_K\}$ be the set of tasks. The model has shared parameters $\theta\in\mathbb R^{d}$ and task-specific parameters $\phi_k\in\mathbb R^{d_k}$ for $T_k$.
Each task draws examples $(x,y_k)$ from a distribution $\mathcal D_k$ and defines a per-example loss $\ell_k(\theta,\phi_k;x,y_k)$. Its population loss is
\begin{equation}
L_k(\theta,\phi_k):=\mathbb E_{(x,y_k)\sim\mathcal D_k}\big[\ell_k(\theta,\phi_k;x,y_k)\big].
\end{equation}
We minimize the standard weighted MTL objective
\begin{equation}
F(\theta,\phi_1,\dots,\phi_K)\;=\;\sum_{k=1}^{K}w_k\,L_k(\theta,\phi_k),
\end{equation}
with nonnegative task weights $w_k$ (default $w_k=1$).
Note that, for simplicity in later sections, we absorb $w_k$ into the per-task gradient estimates. This is permissible since positive scalings do not change cosine signs or the induced conflict graph.
We write $\mathcal L_k(\theta,\phi_k;\mathcal B)$ for the corresponding weighted mini batch loss. %

At step $t$, for any task $k$ that is active we compute stochastic gradients on a mini-batch $\mathcal B_k^{(t)}\subset\mathcal D_k$:
\begin{equation}
g_k^{(t)}:=\nabla_\theta \mathcal L_k(\theta_t,\phi_{k,t};\mathcal B_k^{(t)}),
\qquad
h_k^{(t)}:=\nabla_{\phi_k} \mathcal L_k(\theta_t,\phi_{k,t};\mathcal B_k^{(t)}).
\end{equation}
In our proposed method, we form exponential moving averages (EMA) of per-task gradients within a refresh window to stabilize cosine estimates so that they do not become stale (Sec.~\ref{sec:proposedapproach}) \cite{Polyak1992}.

\subsubsection{Interference Coefficient}
We quantify pairwise interaction \revisedRFour{with the interference coefficient}
\begin{equation}
\rho_{ij}\;=\;-\frac{\langle \tilde g_i,\tilde g_j\rangle}{\|\tilde g_i\|\,\|\tilde g_j\|},
\end{equation}
\revisedRFour{where $\tilde g_i$ and $\tilde g_j$ are the EMA-smoothed gradients at refresh.} Positive $\rho_{ij}$ indicates conflict (negative cosine). $\rho_{ij}\le 0$ indicates alignment or neutrality.

\subsubsection{Conflict Graph}
Fix a tolerance $\tau\in(0,1)$. The conflict graph is
\begin{equation}
G_\tau=(\mathcal T,E_\tau),\qquad
E_\tau=\bigl\{(i,j):\rho_{ij}>\tau\bigr\}.
\end{equation}
Vertices are tasks. An edge between a pair means to not update that pair together.  We will utilize $G_\tau$ for coloring and scheduling in Section \ref{sec:proposedapproach} %

\subsection{Goal}\label{sec:goal}
At training step $t$ we choose an active set $S_t\subseteq\mathcal T$ and update only those tasks:
\begin{equation}
\theta_{t+1}=\theta_t-\eta_t\sum_{k\in S_t} g_k^{(t)},\qquad
\phi_{k,t+1}=
\begin{cases}
\phi_{k,t}-\eta_t h_k^{(t)}, & k\in S_t,\\
\phi_{k,t}, & k\notin S_t.
\end{cases}
\end{equation}
The problem the scheduler addresses is to design the sequence $\{S_t\}_{t=1}^{T}$ \revisedRFour{so that:
(1) every task is visited regularly; and
(2) conflicting tasks seldom appear together.}
We instantiate this via greedy graph coloring in Section \ref{sec:proposedapproach} and analyze the guarantees in Section \ref{sec:theoretical_analysis}.

\section{Proposed Approach}\label{sec:proposedapproach}
\begin{figure*}[!t]
  \centering
  \includegraphics[width=\textwidth]{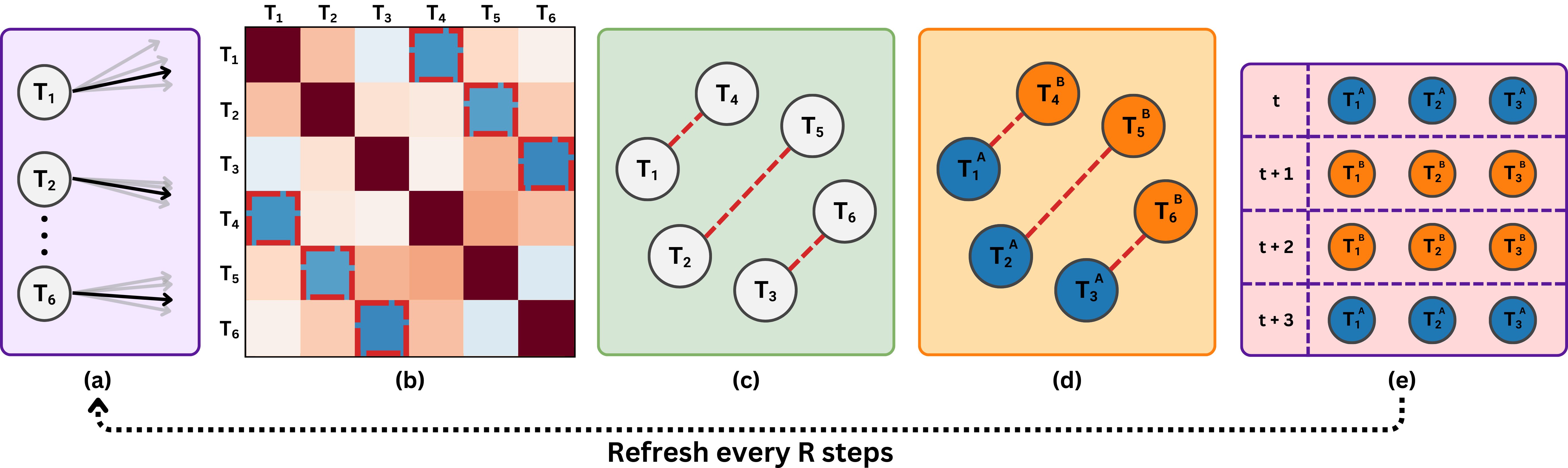}
  \caption{Interference-aware scheduling pipeline: (a) For each task $T_i$ (circles $T_1\ldots T_6$), we smooth recent per-step gradients with an Exponential Moving Average (EMA); (b) From these EMA vectors we compute the pairwise cosine matrix. In the figure, cells outlined with red dashes mark pairs with cosine $<-\tau$. These are flagged as conflicts; (c) We build the conflict graph whose nodes are tasks $T_i$ and whose red dashed edges connect exactly those pairs identified in (b); (d) We apply greedy graph coloring so that no conflict edge lies within a color, producing low-conflict groups. In the example shown, we have two groups: A as blue and B as orange; (e) During training we activate one group per step. After every $R$ steps (here, $R=4$) we 'refresh' and run the pipeline again from step A,
  where we update the EMAs with the latest gradients.
  }
  \label{fig:qavt-arch}
\end{figure*}

We design an interference-aware scheduler that partitions tasks into low-conflict groups and activates exactly one group per optimization step. The procedure consists of four stages: \revisedRFour{(1) estimating pairwise interference}, (2) building and coloring the conflict graph, (3) generating a periodic schedule, and \revisedRFour{(4) updating that schedule as training evolves}. An overview of the scheduler is provided as Algorithm 1 in the Supplementary Materials. A visualization of SON-GOKU is provided in Figure \ref{fig:qavt-arch} alongside a simple summary in the Figure caption. %

\subsection{Estimating Gradient Interference}\label{sec:estimate_gradient_interference}
Throughout this section, $R$ denotes the refresh period. We use $r$ for refresh rounds, $t_r$ for the first training step of refresh round $r$, and $m_r$ for the number of color classes produced in that round. We use $\dsk$ for the sketch width. %
We absorb task weights into per-task losses, so $g_k^{(t)}$ is the gradient of the weighted mini batch loss. Cosine calculations and graph construction are not impacted by applying positive scaling.

At step $t$ and for every task $T_k$ appearing in the current mini-batch we compute a task‑specific stochastic gradient

\begin{equation}\begin{aligned}
g_k^{(t)}=\nabla_\theta \mathcal L_k\!\bigl(\theta_t,\phi_{k,t};\mathcal B_k^{(t)}\bigr),
\end{aligned}
\end{equation}

using an independent sub-batch $\mathcal B_k^{(t)}\subset\mathcal D_k$. We then update an exponential moving average

\begin{equation}\label{eq:ema_def}\begin{aligned}
\tilde g_k^{(t)}=\beta\,\tilde g_k^{(t-1)}+(1-\beta)\,g_k^{(t)},\qquad\beta\in[0,1),
\end{aligned}
\end{equation}

which stabilizes cosine estimates while requiring only two buffers per task (current and previous). \revisedRTwo{To estimate such cosines in practice, we employ low dimensional sketches of the EMA for each task, so the additional memory usage scales well  \cite{Ghashami2016, woodruff2014sketching}.} \revisedRFour{Whenever we refresh the schedule (every $R$ steps) we form the pairwise interference matrix}.
The EMA gradients are used only to build the graph. The parameter updates use the current stochastic gradients.
\begin{equation}\label{eq:rho-neg-cosine}
\begin{aligned}
\rho_{ij}^{(t)} \;=\;
-\frac{\langle\tilde g_i^{(t)},\tilde g_j^{(t)}\rangle}
       {\|\tilde g_i^{(t)}\|\,\|\tilde g_j^{(t)}\|},
\qquad i,j\in\{1,\dots ,K\}.
\end{aligned}
\end{equation}

Computing all $K(K-1)/2$ cosines via the Gram matrix in the $\dsk$-dimensional
sketch space costs $O(K d \dsk + K^2 \dsk)$, namely $O(K d \dsk)$ to form the sketch $M_f$ and $O(K^2 \dsk)$ for the Gram
product, where $\dsk \ll d$ is the sketch width (see Supp. Materials Section~D.5.1).
We also write $h_k^{(t)} = \nabla_{\phi_k}\mathcal L_k\!\bigl(\theta_t,\phi_{k,t};\mathcal B_k^{(t)}\bigr)$ for the gradient with respect to the task-specific parameters $\phi_k$.

\subsection{Conflict Graph Construction}
Given a tolerance $\tau\in(0,1)$, the conflict graph at update round $r$ is

\begin{equation}\label{eq:rho-threshold-graph}
\begin{aligned}
G^{(r)}_\tau=(V,E_\tau^{(r)}),\quad V=\{1,\dots,K\}
E_\tau^{(r)}=\bigl\{(i,j)\;:\;\rho^{(t_r)}_{ij}>\tau\bigr\}.
\end{aligned}
\end{equation}

To clarify, tasks are indexed by integers $1\ldots K$ in Equation \ref{eq:rho-threshold-graph}. Edges connect tasks whose averaged gradients have cosine similarity less than $-\tau$. Intuitively, larger $\tau$ yields a sparser conflict graph, typically fewer colors (larger per-step groups), and more frequent updates per task. Smaller $\tau$ results in a denser graph, more colors (smaller per-step groups), and less frequent updates per task. \revisedRFour{This construction reflects {optimization-time interference}. $G^{(r)}_\tau$ is symmetric and undirected, derived from current gradient geometry to decide which tasks should not be updated together.}

\subsection{Partitioning via Greedy Graph Coloring}\label{sec:partitioning}
We apply the Welsh-Powell largest-first greedy heuristic \cite{Welsh1967, Brelaz-DSATUR} to color 
$G^{(r)}_{\tau}$
and obtain color classes 
$C^{(r)}_{1},\dots ,C^{(r)}_{m_r}$. Classical graph-theory results \cite{West2000, Diestel2017} guarantee the heuristic uses no more than $\Delta+1$ colors, where $\Delta$ is the maximum vertex degree. In practice $\Delta$ is small because many task pairs do not interfere, yielding concise schedules.

\subsection{Schedule Generation and Execution}\label{sec:schedule_generation}
We create a periodic schedule of length $m_r$:

\begin{equation}\begin{aligned}
S_{t} \;=\; C^{(r)}_{\,\bigl(t \bmod m_r\bigr)+1},
\qquad t_r \le t < t_{r+1}=t_r+R.
\end{aligned}
\end{equation}

Each training step activates exactly one color class; over one period every task in that class receives a gradient update, while conflicting tasks (edges in $E_\tau^{(r)}$) are guaranteed not to co-occur.

\subsubsection{Minimum update frequency}\label{sec:min_coverage_constraint}
If the greedy coloring yields a singleton class for a rarely updated task, %
we increase its update frequency by duplicating it only into steps whose active color has no conflict edge to that task.

\subsubsection{Warm-up and Annealing}
We start with $\tau = 1$ (no edges, full simultaneous training) for the first $T_{\mathrm{warm}}$ steps, then logarithmically anneal $\tau$ to a target value $\tau^*$ \cite{goyal2017accurate}. This mitigates noisy gradient signals early in training. \revisedRTwo{Similarly, we can set the refresh period with a smaller $R$ to adapt to changing gradients and increase it as training stabilizes (Supp. Materials Section O).}

\subsection{Time Complexity \revisedRTwo{and Space Complexity}}\label{sec:time_complexity_sec}
\revisedRTwo{Using the sketched implementation described in Supp. Materials Section~D, a single refresh
of the SON-GOKU scheduler has time complexity
$O(K d \dsk + K^2 \dsk)$,
where $\dsk \ll d$ is the sketch width.}
However, unlike many MTL approaches, our scheduler concentrates its extra work in occasional refreshes. \revisedRTwo{This time complexity therefore becomes $O\!\left(\frac{K \dsk(d+K)}{R}\right)$} amortized per training step where $R$ is the refresh period (the number of training steps between conflict-graph rebuilds). \revisedRTwo{For the small, fixed $\dsk$ used in our experiments, this overhead still grows roughly quadratically in $K$ but is independent of $d$ up to the $O(K d \dsk)$ sketching term and shrinks linearly with the refresh period $R$.} \revisedRTwo{Similarly, SON-GOKU's persistent space complexity of $O(K^{2})$ scales with $K$ but not $d$, the number of model parameter dimensions, allowing it to maintain low memory usage even with large backbone models.} We provide a full analysis of the time complexity in Supp. Materials Section D and discuss approaches to reducing time complexity under certain conditions in Supp. Materials Section D.5. See also Supp. Materials Section R for scaling behavior with larger backbones.

\section{Theoretical Analysis}\label{sec:theoretical_analysis}
We discuss some of the main guarantees behind SON-GOKU. %
For a very 
brief overview:
(1) Updating groups of tasks whose gradients are mostly low-conflict (no internal edges) reduces the objective on average and still achieves the usual $1/\sqrt{T}$ convergence rate; (2) Over a refresh window, scheduling several group updates can beat one mixed update that uses all tasks at once; and (3) With a small number of recent gradient measurements per task (via EMA) and a margin separating conflicts, the estimated conflict graph matches the ideal one, giving a short schedule where every task is updated at least once every $\Delta+1$ steps ($\Delta$ is the maximum number of conflicts for any task).
We provide expanded assumptions, definitions, proofs, reasoning, analysis, etc. in Supp. Materials Sections B--I (\revisedRTwo{see also N}, \revisedRThree{\P--}\revisedRTwo{R}).

\subsection{Descent Preservation Within a Low-conflict Group}\label{sec:descent_preservation_main}
If the active set $S_t$ at step $t$ is $\tau$-compatible, then the combined update formed from the weighted shared gradients is a descent direction with a quantitative lower bound:
\begin{equation}
\Bigl\|\sum_{k\in S_t} g_{k,t}\Bigr\|^2
\;\ge\;
\Bigl(1-\tau\,(|S_t|-1)\Bigr)\sum_{k\in S_t}\|g_{k,t}\|^2
\end{equation}

Thus the step cannot flip to ascent whenever $\tau(|S_t|-1)<1$. This is proved by expanding the polarization identity and controlling cross terms under the $\tau$-compatibility condition (see Supp. Materials Section E). Essentially, this means that SON-GOKU's per-step updates are safe when groups are low conflict. The aggregate direction keeps pointing downhill and the cancellation is quantitatively limited by $\tau$ and group size.

\subsection{Nonconvex Convergence at the Standard Rate up to a Small Factor}\label{sec:nonconvex_convergence_at}
Under standard smoothness and noise conditions (see Supp. Materials Section I) and with steps $\eta=c/\sqrt{T}$, SON-GOKU achieves the usual nonconvex SGD rate, with a mild $(1+\tau)$ factor that reflects within-group conflict:
\begin{equation}
\min_{t<T}\mathbb{E}\,\|\nabla F(\theta_t)\|^2
\;\le\;
\frac{2(F_0-F^\star)}{c\sqrt{T}}\,(1+\tau)
\;+\;\frac{cL\sigma^2}{\sqrt{T}}
\end{equation}

When $\tau=0$, the constant matches the classical bound \cite{bottou2018optimization, GhadimiLan2013}; as $\tau\to 1$, it at most doubles, matching the intuition that conflict can cancel up to half of the progress. %
This demonstrates that scheduling does not degrade asymptotic progress. SON-GOKU preserves the $1/\sqrt{T}$ decay of the gradient norm while controlling the constant through the compatibility threshold $\tau$. In other words, we keep the standard rate of SGD and trade a small constant for reduced interference.
We use SGD in the theory to isolate the scheduler. SON-GOKU selects active tasks before the optimizer step and does not reset optimizer states.

\subsection{When Scheduled Groups Outperform a Single Mixed Update}\label{sec:scheduled_groups}
We compare two ways to use the same gradients gathered at a refresh: a scheduled sequence of per-group steps (i.e., the scheduler used in SON-GOKU) versus a single aggregated step. Using a telescoping $L$-smooth bound and evaluating both trajectories at a common linearization (i.e., expanding $F$ at the refresh start $\theta_{t_r}$ and applying the same first-order model with the same step size) the
scheduled bound is never worse and is strictly better when cross-group interaction terms are sufficiently negative (so mixed updates would cancel progress). 

Essentially, when different groups' gradients pull in opposing directions (so adding them together would cancel progress) the scheduler has an advantage. In that case, taking the updates one group at a time is provably better. Our theory guarantees a larger drop in the objective during that refresh than the one-shot step, even though both use the same step size and the same gradients.
Under the PL condition, the scheduled path maintains the usual contraction factor and gains a nonnegative extra decrease term over the window \cite{karimi2016linear}.

\subsection{Exact Recovery of the Population Conflict Graph and Task Partition}\label{sec:exact_recovery} %
We show that, after observing gradients for only a modest number of steps, the scheduler can exactly reconstruct the true conflict relations among tasks by averaging recent gradients (EMA), computing pairwise cosines, thresholding at $-\tau$, and coloring the resulting graph. 
Under a separation margin $\gamma$ around the threshold (tasks are meaningfully different), bounded noise, and bounded drift within each refresh window, the conflict graph estimated from finite data agrees, with high probability, with the ideal population conflict graph $G^\star\tau$ %
(defined from the pairwise cosines of the true mean gradients $\{\mu_i\}_{i=1}^K$ at the start of the refresh window).
Equivalently, when the uniform cosine estimation error is below $\gamma$, we have $\widehat G_\tau = G^\star_\tau$ and the 
resulting grouping recovers the ground-truth task partition.
This explains why the scheduler’s group structure is trustworthy and ties %
the required number of recent gradient measurements per task 
to interpretable quantities such as noise level, margin, and the number of tasks.
For example, an effective sample size of 
$n_{\mathrm{eff}} \;\gtrsim\; \frac{\sigma^2}{m_0^2\,\gamma^2}\,\log\!\bigl(K/\delta\bigr)$
suffices in our analysis.

\subsection{Scheduling Properties with Few Groups and Bounded Staleness}\label{sec:scheduling_properties_main}

Welsh-Powell greedy coloring uses at most $\Delta+1$ colors on a graph whose maximum degree is $\Delta$ \cite{bonamy2018boundingchifractiondelta}. Running the colors in a fixed cycle means each task is updated at least once every $m\le \Delta+1$ steps. Equivalently, no task waits more than $\Delta$ steps between updates (bounded staleness). 

This means that the schedule length is controlled by the worst conflict degree $\Delta$ rather than by the total number of tasks $K$. This results in two important benefits: (1) 
a minimum update-frequency guarantee, since every task receives an update at least once per cycle of length $\le \Delta+1$; and (2) compatibility with standard bounded-delay conditions used in analyses of asynchronous SGD (e.g., \cite{Recht2011Hogwild, lian2015asynchronous, Mitliagkas2016}), with delay parameter at most $\Delta$. When $\Delta \ll K$, we achieve both low interference (few conflicts per step) and low staleness (short update gaps).

\section{Experimental Setup}\label{sec:experimental_setup}
\subsection{Datasets}
We evaluate across six benchmarks spanning vision, multimodal, and time-series \cite{krizhevsky2009learning, silberman2012indoor, arevalo2017gated, vielzeuf2018centralnet}. For each dataset we specify a small set of primary tasks and add positive and negative auxiliaries to stress interference. Architectures are standard backbones (e.g., ResNet-18 for image tasks \cite{he2016deep}, CNN/BiLSTM for time-series) with task-specific heads. Full dataset and task definitions, auxiliary construction, and architecture details (including preprocessing and head designs) are provided in the Supplementary Materials under section J and Table 4. We provide additional experiments with varying backbones in Supp. Materials Section R.

\subsection{Baseline and State-of-the-Art Comparisons}

We compare against loss-weighting (Uniform, GradNorm, AdaTask), multi-objective (MGDA, Nash-MTL, FairGrad), projection/surgery (PCGrad, CAGrad), and fast adaptive weighting (FAMO) \cite{Chen2017GradNorm, yang2023adatask, mgda, Ban2024-NASH, Yu2020PCGrad, Liu2023FAMO, navon2022multi}. We provide short method notes in Supp. Materials Section K and discuss these approaches in Section \ref{sec:related-work}.

\subsection{Scheduler Extension Models}
In addition to standalone models, we also evaluate combinations of the scheduler with existing approaches.
These combinations use SON-GOKU only to select tasks before the optimizer update.

\begin{enumerate}
    \item \textit{SON-GOKU $+$ AdaTask.} Combines our interference-aware task selection with AdaTask's dynamic loss weighting, applying adaptive weights only to scheduler-selected tasks.
    \item \textit{SON-GOKU $+$ GradNorm Warm Start.} Initializes training with GradNorm for stable gradient magnitudes, then transitions to our scheduler after 3 epochs.
    \item \textit{SON-GOKU $+$ PCGrad.} Applied PCGrad's gradient projection specifically to tasks selected by our scheduler, providing fine-grained conflict resolution within $\tau$-compatible groups.
\end{enumerate}

\subsubsection{Single-Step Conflict Estimation}
Here, we set the history length to $H = 1$, so every recoloring step relies on only the most recent mini-batch gradients to estimate interference. Without aggregation over many past steps, the conflict graph should become highly noisy, causing unstable task groupings from one update window to the next. This variant tests the importance of historical conflict statistics in the scheduler.

\section{Results and Discussion}\label{sec:results_and_discussion}

\begin{table*}[t]
  \centering
  \caption{Performance of Evaluated Approaches Across Datasets. \textit{DM} represents Density-Matched ablation variants}
  \label{tab:results}
  \renewcommand{\arraystretch}{1.28}
  \resizebox{\dimexpr\textwidth\relax}{!}{%
  \begin{tabular}{l ccc *{2}{c} *{2}{c} *{3}{c}}
      \toprule
      \multirow{2}{*}{\textbf{Model}} 
        & \multicolumn{3}{c}{\textbf{Accuracy (\%) $\uparrow$}} 
        & \multicolumn{2}{c}{\textbf{F\&B}} 
        & \multicolumn{2}{c}{\textbf{HEALTH}} 
        & \multicolumn{3}{c}{\textbf{NYUv2}} \\
      \cmidrule(lr){2-4} \cmidrule(lr){5-6} \cmidrule(lr){7-8} \cmidrule(lr){9-11}
        & CIFAR-10 & AV-MNIST & MM-IMDb 
        & Acc. (\%) $\uparrow$ & MAE $\downarrow$
        & Acc. (\%) $\uparrow$ & MAE $\downarrow$
        & Angle Error $\downarrow$ & Seg. mIoU $\uparrow$ & Depth RMSE $\downarrow$ \\
      \midrule
      Uniform
        & $55$ \revisedRThree{$\pm 2.2$} & $63$ \revisedRThree{$\pm 1.5$} & $56$ \revisedRThree{$\pm 2.8$}
        & $45$ \revisedRThree{$\pm 2.4$} & $0.57$ \revisedRThree{$\pm 0.030$}
        & $52$ \revisedRThree{$\pm 2.0$} & $0.54$ \revisedRThree{$\pm 0.024$}
        & $21.6$ \revisedRThree{$\pm 0.27$} & $0.059$ \revisedRThree{$\pm 0.003$} & $0.73$ \revisedRThree{$\pm 0.018$} \\
    \addlinespace[0.1em]
      GradNorm
        & $61$ \revisedRThree{$\pm 1.6$} & $65$ \revisedRThree{$\pm 1.1$} & $58$ \revisedRThree{$\pm 2.0$}
        & $47$ \revisedRThree{$\pm 2.3$} & $0.57$ \revisedRThree{$\pm 0.020$}
        & $53$ \revisedRThree{$\pm 2.1$} & $0.52$ \revisedRThree{$\pm 0.019$}
        & $21.4$ \revisedRThree{$\pm 0.23$} & $0.054$ \revisedRThree{$\pm 0.004$} & $0.65$ \revisedRThree{$\pm 0.016$} \\
    \addlinespace[0.1em]
      MGDA
        & $59$ \revisedRThree{$\pm 2.9$} & $62$ \revisedRThree{$\pm 1.7$} & $56$ \revisedRThree{$\pm 3.3$}
        & $44$ \revisedRThree{$\pm 3.0$} & $0.57$ \revisedRThree{$\pm 0.036$}
        & $53$ \revisedRThree{$\pm 2.5$} & $0.53$ \revisedRThree{$\pm 0.030$}
        & $21.8$ \revisedRThree{$\pm 0.33$} & $0.063$ \revisedRThree{$\pm 0.005$} & $0.75$ \revisedRThree{$\pm 0.024$} \\
    \addlinespace[0.1em]
    \midrule
      PCGrad
        & $61$ \revisedRThree{$\pm 1.9$} & $65$ \revisedRThree{$\pm 1.3$} & $58$ \revisedRThree{$\pm 2.3$}
        & $50$ \revisedRThree{$\pm 2.1$} & $0.55$ \revisedRThree{$\pm 0.024$}
        & $58$ \revisedRThree{$\pm 2.0$} & $0.48$ \revisedRThree{$\pm 0.021$}
        & $20.9$ \revisedRThree{$\pm 0.24$} & $0.070$ \revisedRThree{$\pm 0.004$} & $0.69$ \revisedRThree{$\pm 0.013$} \\
    \addlinespace[0.1em]
      CAGrad
        & $59$ \revisedRThree{$\pm 2.0$} & $62$ \revisedRThree{$\pm 1.1$} & $57$ \revisedRThree{$\pm 2.5$}
        & $46$ \revisedRThree{$\pm 2.5$} & $0.58$ \revisedRThree{$\pm 0.031$}
        & $53$ \revisedRThree{$\pm 1.9$} & $0.52$ \revisedRThree{$\pm 0.024$}
        & $21.9$ \revisedRThree{$\pm 0.29$} & $0.065$ \revisedRThree{$\pm 0.004$} & $0.73$ \revisedRThree{$\pm 0.018$} \\
    \addlinespace[0.1em]
      AdaTask
        & $63$ \revisedRThree{$\pm 1.5$} & $67$ \revisedRThree{$\pm 0.9$} & $59$ \revisedRThree{$\pm 1.9$}
        & $47$ \revisedRThree{$\pm 1.9$} & $0.59$ \revisedRThree{$\pm 0.026$}
        & $55$ \revisedRThree{$\pm 2.2$} & $0.52$ \revisedRThree{$\pm 0.024$}
        & $20.3$ \revisedRThree{$\pm 0.23$} & $0.069$ \revisedRThree{$\pm 0.004$} & $0.65$ \revisedRThree{$\pm 0.015$} \\
    \addlinespace[0.1em]
      FAMO
        & $64$ \revisedRThree{$\pm 1.2$} & $70$ \revisedRThree{$\pm 1.0$} & $61$ \revisedRThree{$\pm 1.6$}
        & $52$ \revisedRThree{$\pm 2.0$} & $0.53$ \revisedRThree{$\pm 0.021$}
        & $60$ \revisedRThree{$\pm 1.8$} & $0.49$ \revisedRThree{$\pm 0.018$}
        & $19.9$ \revisedRThree{$\pm 0.19$} & $0.074$ \revisedRThree{$\pm 0.003$} & $0.63$ \revisedRThree{$\pm 0.012$} \\
    \addlinespace[0.1em]
      FairGrad
        & $62$ \revisedRThree{$\pm 1.8$} & $66$ \revisedRThree{$\pm 1.3$} & $59$ \revisedRThree{$\pm 2.5$}
        & $52$ \revisedRThree{$\pm 2.5$} & $0.54$ \revisedRThree{$\pm 0.026$}
        & $60$ \revisedRThree{$\pm 2.0$} & $0.47$ \revisedRThree{$\pm 0.022$}
        & $20.7$ \revisedRThree{$\pm 0.27$} & $0.072$ \revisedRThree{$\pm 0.004$} & $0.67$ \revisedRThree{$\pm 0.015$} \\
    \addlinespace[0.1em]
      Nash-MTL
        & $63$ \revisedRThree{$\pm 1.9$} & $66$ \revisedRThree{$\pm 1.2$} & $60$ \revisedRThree{$\pm 2.1$}
        & $52$ \revisedRThree{$\pm 2.3$} & $0.54$ \revisedRThree{$\pm 0.024$}
        & $60$ \revisedRThree{$\pm 2.3$} & $0.47$ \revisedRThree{$\pm 0.023$}
        & $20.6$ \revisedRThree{$\pm 0.24$} & $0.073$ \revisedRThree{$\pm 0.004$} & $0.67$ \revisedRThree{$\pm 0.013$} \\
    \midrule
    \addlinespace[0.1em]
      Static One-Shot
        & $61$ \revisedRThree{$\pm 2.0$} & $66$ \revisedRThree{$\pm 1.1$} & $58$ \revisedRThree{$\pm 2.6$}
        & $48$ \revisedRThree{$\pm 2.3$} & $0.56$ \revisedRThree{$\pm 0.027$}
        & $54$ \revisedRThree{$\pm 2.1$} & $0.51$ \revisedRThree{$\pm 0.025$}
        & $20.5$ \revisedRThree{$\pm 0.25$} & $0.071$ \revisedRThree{$\pm 0.004$} & $0.65$ \revisedRThree{$\pm 0.016$} \\
    \addlinespace[0.1em]
      Single-Step
        & $40$ \revisedRThree{$\pm 4.2$} & $59$ \revisedRThree{$\pm 2.4$} & $20$ \revisedRThree{$\pm 5.4$}
        & $42$ \revisedRThree{$\pm 3.9$} & $0.60$ \revisedRThree{$\pm 0.041$}
        & $47$ \revisedRThree{$\pm 3.5$} & $0.55$ \revisedRThree{$\pm 0.034$}
        & $26.4$ \revisedRThree{$\pm 0.55$} & $0.042$ \revisedRThree{$\pm 0.006$} & $0.81$ \revisedRThree{$\pm 0.029$} \\
    \addlinespace[0.1em]
      SON-GOKU (Threshold, DM)
        & 63 & 68 & 59
        & 49 & 0.55
        & 56 & 0.51
        & 20.6 & 0.071 & 0.61 \\
    \addlinespace[0.1em]
      SON-GOKU (kNN-Symm.)
        & 60 & 65 & 55
        & 46 & 0.57
        & 52 & 0.53
        & 22.1 & 0.066 & 0.70 \\
    \addlinespace[0.1em]
      SON-GOKU (kNN-Symm., DM)
        & 61 & 66 & 57
        & 47 & 0.56
        & 54 & 0.52
        & 21.4 & 0.068 & 0.66 \\
    \addlinespace[0.1em]
      SON-GOKU (Signed-only)
        & 56 & 63 & 52
        & 43 & 0.60
        & 50 & 0.56
        & 24.0 & 0.053 & 0.76 \\
    \addlinespace[0.1em]
      SON-GOKU (Signed-only, DM)
        & 58 & 64 & 54
        & 45 & 0.59
        & 52 & 0.54
        & 23.0 & 0.056 & 0.73 \\
    \addlinespace[0.1em]
      SON-GOKU (Quantile)
        & 64 & 68 & 60
        & 50 & 0.54
        & 57 & 0.50
        & 20.3 & 0.072 & 0.60 \\
    \addlinespace[0.1em]
      SON-GOKU (Quantile, DM)
        & 65 & 69 & 61
        & 51 & 0.53
        & 58 & 0.50
        & 20.0 & 0.072 & \textbf{0.59} \\
    \midrule
    \addlinespace[0.1em]
      SON-GOKU $+$ GradNorm
        & $62$ \revisedRThree{$\pm 1.4$} & $69$ \revisedRThree{$\pm 1.0$} & $59$ \revisedRThree{$\pm 1.7$}
        & $51$ \revisedRThree{$\pm 1.8$} & $0.53$ \revisedRThree{$\pm 0.022$}
        & $59$ \revisedRThree{$\pm 1.7$} & $0.49$ \revisedRThree{$\pm 0.018$}
        & $\mathbf{19.6}$ \revisedRThree{$\pm 0.19$} & $0.073$ \revisedRThree{$\pm 0.003$} & $0.64$ \revisedRThree{$\pm 0.011$} \\
    \addlinespace[0.1em]
      SON-GOKU $+$ AdaTask
        & $\mathbf{67}$ \revisedRThree{$\pm 1.2$} & $\mathbf{71}$ \revisedRThree{$\pm 0.9$} & $\mathbf{63}$ \revisedRThree{$\pm 1.6$}
        & $52$ \revisedRThree{$\pm 1.7$} & $0.53$ \revisedRThree{$\pm 0.021$}
        & $59$ \revisedRThree{$\pm 1.8$} & $0.48$ \revisedRThree{$\pm 0.017$}
        & $20.1$ \revisedRThree{$\pm 0.20$} & $0.068$ \revisedRThree{$\pm 0.004$} & $0.67$ \revisedRThree{$\pm 0.013$} \\
    \addlinespace[0.1em]
      SON-GOKU $+$ PCGrad
        & $65$ \revisedRThree{$\pm 1.3$} & $70$ \revisedRThree{$\pm 0.9$} & $60$ \revisedRThree{$\pm 1.8$}
        & $\mathbf{54}$ \revisedRThree{$\pm 2.0$} & $\mathbf{0.52}$ \revisedRThree{$\pm 0.024$}
        & $\mathbf{62}$ \revisedRThree{$\pm 1.6$} & $\mathbf{0.45}$ \revisedRThree{$\pm 0.020$}
        & $19.7$ \revisedRThree{$\pm 0.18$} & $\mathbf{0.076}$ \revisedRThree{$\pm 0.003$} & $0.62$ \revisedRThree{$\pm 0.010$} \\
    \addlinespace[0.1em]
       SON-GOKU
        & $65$ \revisedRThree{$\pm 1.5$} & $69$ \revisedRThree{$\pm 1.0$} & $61$ \revisedRThree{$\pm 1.8$}
        & $51$ \revisedRThree{$\pm 1.9$} & $0.53$ \revisedRThree{$\pm 0.023$}
        & $58$ \revisedRThree{$\pm 1.7$} & $0.50$ \revisedRThree{$\pm 0.018$}
        & $19.8$ \revisedRThree{$\pm 0.20$} & $0.073$ \revisedRThree{$\pm 0.004$} & $\mathbf{0.59}$ \revisedRThree{$\pm 0.012$} \\
      \bottomrule
    \end{tabular}%
}
\end{table*}

Results for all models across every experiment are depicted in Table \ref{tab:results}. \revisedRThree{All metrics are held-out {test} results under identical training setups and architectures.} Across ten metrics on six datasets, our conflict-aware schedulers consistently match or exceed all baseline methods.

\subsection{Overall Performance Improvements}
Overall, the conflict-aware approaches improve over the uniform baseline by 10\%-20\% on CIFAR-10 and by 7\% on MM-IMDb, indicating that grouping tasks according to measured interference is more effective than treating all tasks equally at every update.
On NYUv2, we see similar improvements across all the metrics. These results suggest that the scheduler's graph coloring cleanly separates high-conflict tasks, preserving the projection or LR-balancing advantages (stemming from PCGrad’s gradient projection and AdaTask’s learning‑rate adaptation, respectively) while removing residual interference \revisedRFour{(see Supp. Materials Section S for grouping patterns at training time and more analyses)}. \revisedRThree{As we evaluated across diverse tasks and datasets, our results also demonstrate clear improvements in generalization.}

\subsection{Ablation Study on Scheduler Design}\label{sec:ablation_results}
\revisedROne{
We evaluate nine controlled ablations of six types: (i) \textbf{Static One-Shot Coloring}, which runs greedy graph coloring once at the start of training and then freezes the groups, testing dependence on \emph{dynamic} recoloring as gradients change; (ii) \textbf{Single-Step Conflict Estimation}, which sets the history length to $H=1$ so each recoloring uses only the most recent mini batch, testing the importance of averaging conflict statistics over time; (iii) \textbf{Threshold Graph (baseline)}, which connects tasks $i$ and $j$ when the smoothed cosine $\hat{s}_{ij}(t)$ falls below a global threshold $-\tau(t)$; (iv) \textbf{kNN-Symmetric Graph}, which connects each task to its $m$ most conflicting neighbors and then symmetrizes the edges, enforcing roughly fixed degree per task and comparing local degree control against the global threshold rule; (v) \textbf{Signed-Only Graph}, which adds an edge only if $\hat{s}_{ij}(t) < 0$, yielding a very sparse graph and ignoring moderate (but potentially harmful) conflicts; and (vi) \textbf{Quantile Threshold Graph}, which at each refresh sets $\tau(t)$ so that only the worst $p\%$ of cosine values are treated as conflicting, keeping edge density approximately stable and testing an adaptive cutoff versus a fixed global threshold.}
\revisedROne{We evaluate each graph rule under two settings. In the \emph{fixed $\tau$} setting, all rules share the same $\tau(t)$ schedule used in the main experiments. In the \emph{density-matched} setting, we adjust the hyperparameters of each rule so that all graphs have approximately the same edge density at each refresh. This isolates the effect of {which} pairs are marked as conflicting, rather than how many edges are present. We go into much further detail regarding the ablation in Supp. Materials Section K.3.}

\revisedROne{
These ablations directly test the assumptions behind SON-GOKU. Static One-Shot, which freezes groups, consistently underperforms the full scheduler on most metrics, indicating that task relations change enough during training that dynamic recoloring is needed to maintain $\tau$-compatibility as gradients drift (Sections~\ref{sec:descent_preservation_main}--\ref{sec:nonconvex_convergence_at}). Single-Step, which uses $H = 1$, is clearly worse across datasets, matching our claim that batch cosines are too noisy \cite{keskar2017on}. Instead, averaging conflict statistics over short history windows provides the clean information needed for accurate graph recovery (Section~\ref{sec:exact_recovery}). Among graph constructions, simple threshold and quantile rules (and their density-matched variants) perform similarly well, suggesting that any approach that reliably isolates the worst conflicting pairs is sufficient. In contrast, Signed-Only and kNN-Symmetric, which ignore conflict magnitude or have purely local degree control, degrade performance more noticeably, especially on NYUv2 and the tabular benchmarks. Overall, the best performing configurations are precisely those that match the descent and recovery conditions analyzed in Sections~\ref{sec:descent_preservation_main}--\ref{sec:nonconvex_convergence_at} and~\ref{sec:exact_recovery}.}

\subsection{Additional Analysis}
\subsubsection{Optimizer{-}Task Alignment}
Interestingly, we observe that AdaTask-based approaches tend to be the best on classification tasks (CIFAR-10, AV-MNIST, MM-IMDb) while PCGrad-based approaches tend to be the best on tasks that model regression (NYUv2).

We believe that this stems from unique differences in the features of classification and regression-based models. For example, cross-entropy gradients near decision boundaries tend to be bursty and high in variance \cite{Shrivastava2016OHEM, Lin2017FocalLoss, Hoffer2017TrainLonger, smith2017bayesian}. By scaling each task's step size according to its running gradient norm, AdaTask smooths out these spikes.

On the other hand, we believe that PCGrad under the scheduler performs particularly well on regression and dense-prediction tasks as their tasks tend to generate smooth, large-magnitude gradients whose directions change gradually. PCGrad removes only the small component of the gradient that conflicts across tasks, preserving the main descent direction while reducing interference.

\subsubsection{Synergy Between Scheduling and Baselines}
We believe that the superior results found in the combinations of the scheduler and baseline models can be traced to the way scheduling and optimization reinforce one another.

First, greedy graph coloring partitions tasks into $\tau$-compatible groups, segregating tasks with highly divergent gradients. This yields a guaranteed lower bound on descent (Proposition 6 under Supplementary Materials), directly improving optimization efficiency.

Within each low-conflict group, the optimizer can do its job under more ideal conditions. PCGrad can remove the remaining minor conflicting components, preserving the majority of the descent direction. AdaTask can adjust each task's learning rate without being impacted by large adversarial gradients.

This $\Delta+1$ color bound ensures that every task is scheduled at least once per period. This prevents tasks from being essentially starved of updates. 

Finally, by computing interference over a window, the scheduler smooths out gradient fluctuations \cite{mandt2017stochastic}. This prevents the erratic schedule changes that projection-only grouping methods have been shown to face \cite{Yu2020PCGrad, Shi2023Recon, Zhang2024Proactive}, thereby better stabilizing convergence.

\revisedRThree{
\subsubsection{Optimization Structure and Held-out Performance}\label{sec:opt_to_gen}
While our guarantees in Section~\ref{sec:theoretical_analysis} and Supp. Materials Sections~B--F are stated in optimization terms, they help explain the held out gains by increasing gradient coherence and limiting destructive interference. Section~\ref{sec:descent_preservation_main} shows that the aggregated group gradient remains aligned with descent and that intra-group gradient conflict is explicitly limited by $\tau$ and $|S_t|$. Section~\ref{sec:scheduled_groups} then compares two ways to apply the same gradients during a refresh, either a single mixed update or a scheduled sequence of group updates. Together, these analyses imply that each step in SON-GOKU provides more informative signals and less interference, or, equivalently, a higher gradient-to-noise ratio \cite{sun2023unleashing, Fan2023MaxGNR, mccandlish2018empirical, smith2017bayesian, Mandt2016}. Building on this, Section~\ref{sec:exact_recovery} shows that SON-GOKU's estimated conflict graph recovers the population structure with high probability, so the schedule repeatedly updates clusters of related tasks rather than conflicting tasks. By enforcing positive affinity within groups, SON-GOKU is able to train related tasks together. This enables effective sharing of model parameters across different tasks, reducing the complexity of the model and increasing sample efficiency \cite{Caruana1997, Argyriou2007MultiTaskFeature, 9336293}. With this alongside a high gradient-to-noise signal ratio, SON-GOKU can improve performance across many different datasets, domains, and distributions and can perform well even under non-ideal conditions (e.g., noisy labels, class or task imbalance, distribution shift, etc.) \cite{michalkiewicz2023domain}. Our ablation results (Table \ref{tab:results}) demonstrate that variants without dynamic recoloring or history averaging perform worse, indicating accurate and low-conflict grouping is essential.}

\subsection{Speed and Tradeoffs}\label{sec:speed_tradeoffs}
\begin{table*}[t]
  \centering
  \caption{Wall-clock time (seconds $\pm$ standard deviation) vs. number of tasks \(K\).}
  \label{tab:main_time}
  \renewcommand{\arraystretch}{1.28}
  \resizebox{\textwidth}{!}{%
  \begin{tabular}{lcccc}
    \toprule
    \textbf{Method (R if applicable)} & \textbf{K=3} & \textbf{K=6} & \textbf{K=16} & \textbf{K=40} \\
    \midrule
    Uniform                               & 0.2656 $\pm$ 0.1201 & 0.3240 $\pm$ 0.0629 & 0.3798 $\pm$ 0.1050 & 0.4054 $\pm$ 0.1190 \\
    GradNorm                              & 5.4714 $\pm$ 0.7137 & 5.1201 $\pm$ 0.6112 & 4.9042 $\pm$ 0.5869 & 4.7372 $\pm$ 0.9286 \\
    AdaTask                                & 2.1816 $\pm$ 0.0934 & 2.1032 $\pm$ 0.1012 & 2.2853 $\pm$ 0.0718 & 2.2278 $\pm$ 0.1370 \\
    PCGrad                                 & 3.6212 $\pm$ 0.3517 & 23.1266 $\pm$ 0.8773 & 176.7566 $\pm$ 2.8171 & 1127.1337 $\pm$ 34.2603 \\
    MGDA                                  & 97.1081 $\pm$ 5.4645 & 121.4371 $\pm$ 9.0923 & 132.4913 $\pm$ 3.1752 & 134.0878 $\pm$ 2.2621 \\
    FAMO                   & 2.0725 $\pm$ 0.2073 & 1.9980 $\pm$ 0.1998 & 2.1710 $\pm$ 0.2171 & 2.1164 $\pm$ 0.2116 \\
    FairGrad               & 3.8020 $\pm$ 0.5703 & 15.2079 $\pm$ 2.2812 & 108.1450 $\pm$ 16.2218 & 675.9065 $\pm$ 101.3860 \\
    Nash-MTL               & 5.7030 $\pm$ 1.1406 & 22.8118 $\pm$ 4.5624 & 162.2176 $\pm$ 32.4435 & 1013.8598 $\pm$ 202.7720 \\
    \addlinespace[0.1em]
    \midrule
    SON-GOKU ($R=32$)                 & 1.9896 $\pm$ 0.3651 & 3.3202 $\pm$ 0.5745 & 6.0897 $\pm$ 0.9425 & 12.1432 $\pm$ 1.2044 \\
    SON-GOKU $+$ AdaTask ($R=32$)          & 3.7718 $\pm$ 0.9654 & 5.0511 $\pm$ 0.6531 & 7.5903 $\pm$ 1.1920 & 14.5182 $\pm$ 2.0660 \\
    SON-GOKU $+$ GradNorm ($R=32$)         & 7.0202 $\pm$ 1.0711 & 8.1661 $\pm$ 0.9355 & 10.7227 $\pm$ 2.2088 & 16.5760 $\pm$ 1.8418 \\
    SON-GOKU $+$ PCGrad ($R=32$)           & 1.9834 $\pm$ 0.3586 & 3.4971 $\pm$ 0.3840 & 6.1395 $\pm$ 0.9425 & 10.9097 $\pm$ 1.5263 \\
    \bottomrule
  \end{tabular}}
\end{table*}

SON-GOKU has a time complexity of $O\!\big(K \dsk (d+K)/R\big)$ (Section~4.5) amortized per training step 
(Section \ref{sec:time_complexity_sec}). Table~\ref{tab:main_time} shows near-linear growth over this range of $K$ at $R{=}32$, reflecting sparsity in the graphs and batched cosine computation. SON-GOKU's time rises from around 2 seconds ($K=3$) to 12 seconds ($K=40$), remaining far below methods that perform heavy conflict handling. For example, PCGrad, FairGrad, and Nash-MTL increase steeply with $K$. In contrast, FAMO and AdaTask are among the fastest and largely flat with $K$, as expected from their constant overhead.

\revisedRTwo{SON-GOKU is also memory efficient, with an only incremental memory footprint that scales with the number of tasks \(K\), not the parameter dimension \(d\). The scheduler's peak memory during a refresh step is \(O(K^{2}{+}K\dsk)\) and the persistent state between refreshes is \(O(K^{2})\) (see Supp. Materials Section~N for further theoretical and experimental analysis). By contrast, methods that retain \(K\) full gradients require \(O(Kd)\) additional memory. This implies that, on larger backbones (high \(d\)), SON-GOKU's memory overhead is modest and grows mainly with the task count \(K\), rather than with model size.}

These contrasts demonstrate the 
tradeoffs between speed and fidelity to task interference. Faster methods like FAMO minimize overhead, while methods that model conflicts can improve accuracy. These tradeoffs have to be assessed on a case-by-case basis, based on values
that factor into each approach's time complexity and the importance of training speed versus performance.

\section{Conclusion}

We introduced SON-GOKU, an interference-aware scheduler that estimates cross-task alignment, builds a sparse conflict graph, and greedily colors it to activate one low-conflict group per step. 
Formally, we provide rigorous theoretical guarantees that justify the design and effectiveness of the scheduler. 
Empirically, across six benchmarks, SON-GOKU improves over strong baselines and recent approaches. It complements optimizers like PCGrad and AdaTask, indicating that scheduling and gradient shaping are synergistic. By modeling task interactions with a conflict graph and schedule, SON-GOKU offers a simple, scalable, and theory-backed mechanism for robust multitask training.

\section*{Acknowledgements}
This work was supported in part by the Lambda Research Grant, which provided GPU cloud compute resources through Lambda Cloud. We also gratefully acknowledge the Google TPU Research Cloud (TRC) program for providing Cloud TPU compute resources used in our experiments.

\bibliographystyle{splncs04}
\bibliography{main, appendix, additional_20, additional_second}

@article{Baxter2000,
author = {Baxter, Jonathan},
title = {A model of inductive bias learning},
year = {2000},
issue_date = {February 2000},
publisher = {AI Access Foundation},
address = {El Segundo, CA, USA},
volume = {12},
number = {1},
issn = {1076-9757},
journal = {J. Artif. Int. Res.},
month = mar,
pages = {149–198},
numpages = {50}
}

@inproceedings{mgda,
author = {Sener, Ozan and Koltun, Vladlen},
title = {Multi-task learning as multi-objective optimization},
year = {2018},
publisher = {Curran Associates Inc.},
address = {Red Hook, NY, USA},
booktitle = {Proceedings of the 32nd International Conference on Neural Information Processing Systems},
pages = {525–536},
numpages = {12},
location = {Montr\'{e}al, Canada},
series = {NIPS'18}
}

@article{navon2022multi,
  title={Multi-task learning as a bargaining game},
  author={Navon, Aviv and Shamsian, Aviv and Achituve, Idan and Maron, Haggai and Kawaguchi, Kenji and Chechik, Gal and Fetaya, Ethan},
  journal={arXiv preprint arXiv:2202.01017},
  year={2022}
}

@inproceedings{Ban2024-NASH,
author = {Ban, Hao and Ji, Kaiyi},
title = {Fair resource allocation in multi-task learning},
year = {2024},
publisher = {JMLR.org},
booktitle = {Proceedings of the 41st International Conference on Machine Learning},
articleno = {109},
numpages = {17},
location = {Vienna, Austria},
series = {ICML'24}
}

@article{lin2019pareto,
  title={Pareto multi-task learning},
  author={Lin, Xi and Zhen, Hui-Ling and Li, Zhenhua and Zhang, Qing-Fu and Kwong, Sam},
  journal={Advances in neural information processing systems},
  volume={32},
  year={2019}
}

@inproceedings{karimi2016linear,
  title={Linear convergence of gradient and proximal-gradient methods under the polyak-{\l}ojasiewicz condition},
  author={Karimi, Hamed and Nutini, Julie and Schmidt, Mark},
  booktitle={Joint European conference on machine learning and knowledge discovery in databases},
  pages={795--811},
  year={2016},
  organization={Springer}
}

@article{goyal2017accurate,
  title={Accurate, large minibatch sgd: Training imagenet in 1 hour},
  author={Goyal, Priya and Doll{\'a}r, Piotr and Girshick, Ross and Noordhuis, Pieter and Wesolowski, Lukasz and Kyrola, Aapo and Tulloch, Andrew and Jia, Yangqing and He, Kaiming},
  journal={arXiv preprint arXiv:1706.02677},
  year={2017}
}

@article{woodruff2014sketching,
  title={Sketching as a tool for numerical linear algebra},
  author={Woodruff, David P and others},
  journal={Foundations and Trends{\textregistered} in Theoretical Computer Science},
  volume={10},
  number={1--2},
  pages={1--157},
  year={2014},
  publisher={Now Publishers, Inc.}
}

@inproceedings{silberman2012indoor,
  title={Indoor segmentation and support inference from rgbd images},
  author={Silberman, Nathan and Hoiem, Derek and Kohli, Pushmeet and Fergus, Rob},
  booktitle={European conference on computer vision},
  pages={746--760},
  year={2012},
  organization={Springer}
}

@article{arevalo2017gated,
  title={Gated multimodal units for information fusion},
  author={Arevalo, John and Solorio, Thamar and Montes-y-G{\'o}mez, Manuel and Gonz{\'a}lez, Fabio A},
  journal={arXiv preprint arXiv:1702.01992},
  year={2017}
}

@inproceedings{vielzeuf2018centralnet,
  title={Centralnet: a multilayer approach for multimodal fusion},
  author={Vielzeuf, Valentin and Lechervy, Alexis and Pateux, St{\'e}phane and Jurie, Fr{\'e}d{\'e}ric},
  booktitle={Proceedings of the European conference on computer vision (ECCV) workshops},
  pages={0--0},
  year={2018}
}

@inproceedings{he2016deep,
  title={Deep residual learning for image recognition},
  author={He, Kaiming and Zhang, Xiangyu and Ren, Shaoqing and Sun, Jian},
  booktitle={Proceedings of the IEEE conference on computer vision and pattern recognition},
  pages={770--778},
  year={2016}
}

@article{smith2017bayesian,
  title={A bayesian perspective on generalization and stochastic gradient descent},
  author={Smith, Samuel L and Le, Quoc V},
  journal={arXiv preprint arXiv:1710.06451},
  year={2017}
}

@inproceedings{
keskar2017on,
title={On Large-Batch Training for Deep Learning: Generalization Gap and Sharp Minima},
author={Nitish Shirish Keskar and Dheevatsa Mudigere and Jorge Nocedal and Mikhail Smelyanskiy and Ping Tak Peter Tang},
booktitle={International Conference on Learning Representations},
year={2017},
url={https://openreview.net/forum?id=H1oyRlYgg}
}

@article{mandt2017stochastic,
  title={Stochastic gradient descent as approximate bayesian inference},
  author={Mandt, Stephan and Hoffman, Matthew D and Blei, David M},
  journal={Journal of Machine Learning Research},
  volume={18},
  number={134},
  pages={1--35},
  year={2017}
}

@article{Polyak1992,
author = {Polyak, B. T. and Juditsky, A. B.},
title = {Acceleration of Stochastic Approximation by Averaging},
journal = {SIAM Journal on Control and Optimization},
volume = {30},
number = {4},
pages = {838-855},
year = {1992},
doi = {10.1137/0330046},
URL = { 
        https://doi.org/10.1137/0330046

},
eprint = { 
        https://doi.org/10.1137/0330046

},
urldate = {2026-06-27},
note = {Accessed 27 June 2026}
}

@article{Brelaz-DSATUR,
author = {Br\'{e}laz, Daniel},
title = {New methods to color the vertices of a graph},
year = {1979},
issue_date = {April 1979},
publisher = {Association for Computing Machinery},
address = {New York, NY, USA},
volume = {22},
number = {4},
issn = {0001-0782},
url = {https://doi.org/10.1145/359094.359101},
doi = {10.1145/359094.359101},
journal = {Commun. ACM},
month = apr,
pages = {251–256},
numpages = {6},
urldate = {2026-06-27},
note = {Accessed 27 June 2026}
}

@inproceedings{Mitliagkas2016,
author = {Mitliagkas, Ioannis and Zhang, Ce and Hadjis, Stefan and R\'{e}, Christopher},
title = {Asynchrony begets momentum, with an application to deep learning},
year = {2016},
publisher = {IEEE Press},
url = {https://doi.org/10.1109/ALLERTON.2016.7852343},
doi = {10.1109/ALLERTON.2016.7852343},
booktitle = {2016 54th Annual Allerton Conference on Communication, Control, and Computing (Allerton)},
pages = {997–1004},
numpages = {8},
location = {Monticello, IL, USA},
urldate = {2026-06-27},
note = {Accessed 27 June 2026}
}

@inproceedings{Mandt2016,
author = {Mandt, Stephan and Hoffman, Matthew D. and Blei, David M.},
title = {A variational analysis of stochastic gradient algorithms},
year = {2016},
publisher = {JMLR.org},
booktitle = {Proceedings of the 33rd International Conference on International Conference on Machine Learning - Volume 48},
pages = {354–363},
numpages = {10},
location = {New York, NY, USA},
series = {ICML'16}
}

@String(CVPR= {IEEE Conf. Comput. Vis. Pattern Recog.})

@String(ICCV= {Int. Conf. Comput. Vis.})

@String(ECCV= {Eur. Conf. Comput. Vis.})

@String(NIPS= {Adv. Neural Inform. Process. Syst.})

@String(ICLR = {Int. Conf. Learn. Represent.})

@String(AAAI = {AAAI})

@String(CVPR  = {CVPR})

@String(ICCV  = {ICCV})

@String(ECCV  = {ECCV})

@String(NIPS  = {NeurIPS})

@String(ICLR  = {ICLR})

@article{bottou2018optimization,
  title={Optimization methods for large-scale machine learning},
  author={Bottou, L{\'e}on and Curtis, Frank E and Nocedal, Jorge},
  journal={SIAM review},
  volume={60},
  number={2},
  pages={223--311},
  year={2018},
  publisher={SIAM}
}

@article{Ghashami2016,
  author  = {Mina Ghashami and Edo Liberty and Jeff M. Phillips and David P. Woodruff},
  title   = {Frequent Directions: Simple and Deterministic Matrix Sketching},
  journal = {SIAM Journal on Computing},
  year    = {2016},
  volume  = {45},
  number  = {5},
  pages   = {1762--1792},
  doi     = {10.1137/15M1009718}
}

@misc{bonamy2018boundingchifractiondelta,
      title={Bounding $\chi$ by a fraction of $\Delta$ for graphs without large cliques}, 
      author={Marthe Bonamy and Tom Kelly and Peter Nelson and Luke Postle},
      year={2018},
      eprint={1803.01051},
      archivePrefix={arXiv},
      primaryClass={math.CO},
      url={https://arxiv.org/abs/1803.01051}, 
      urldate = {2025-07-31},
      note = {Accessed July 31 2025},
}

@article{Welsh1967,
    author = {Welsh, D. J. A. and Powell, M. B.},
    title = {An upper bound for the chromatic number of a graph and its application to timetabling problems},
    journal = {The Computer Journal},
    volume = {10},
    number = {1},
    pages = {85-86},
    year = {1967},
    month = {01},
    issn = {0010-4620},
    doi = {10.1093/comjnl/10.1.85},
    url = {https://doi.org/10.1093/comjnl/10.1.85},
    urldate = {2025-08-24},
    note = {Accessed August 24 2025},
    eprint = {https://academic.oup.com/comjnl/article-pdf/10/1/85/1069035/100085.pdf},
}

@book{West2000,
  title     = {Introduction to Graph Theory},
  author    = {West, Douglas B.},
  edition   = {2nd},
  publisher = {Prentice Hall},
  year      = {2000},
  isbn      = {978-0130144003}
}

@article{GhadimiLan2013,
  title        = {Stochastic First- and Zeroth-Order Methods for Nonconvex Stochastic Programming},
  author       = {Ghadimi, Saeed and Lan, Guanghui},
  journal      = {SIAM Journal on Optimization},
  volume       = {23},
  number       = {4},
  pages        = {2341--2368},
  year         = {2013},
  doi          = {10.1137/120880811}
}

@inproceedings{Recht2011Hogwild,
  title        = {HOGWILD!: A Lock-Free Approach to Parallelizing Stochastic Gradient Descent},
  author       = {Niu, Feng and Recht, Benjamin and Ré, Christopher and Wright, Stephen J.},
  booktitle    = {Advances in Neural Information Processing Systems},
  volume       = {24},
  pages        = {693--701},
  year         = {2011}
}

@book{Diestel2017,
  title        = {Graph Theory},
  author       = {Diestel, Reinhard},
  edition      = {5th},
  publisher    = {Springer},
  year         = {2017},
  isbn         = {978-3-662-53622-3}
}

@article{Ruder2017Overview,
  title        = {An Overview of Multi-Task Learning in Deep Neural Networks},
  author       = {Ruder, Sebastian},
  journal      = {arXiv preprint arXiv:1706.05098},
  year         = {2017},
  urldate = {2025-12-04},
  note = {Accessed December 4 2025},
  url          = {https://arxiv.org/abs/1706.05098}
}

@inproceedings{Standley2020TaskGrouping,
  title        = {Which Tasks Should Be Learned Together in Multi-Task Learning?},
  author       = {Standley, Trevor and Zamir, Amir R. and Chen, Dawn and Guibas, Leonidas and Malik, Jitendra and Savarese, Silvio},
  booktitle    = {Proceedings of the 37th International Conference on Machine Learning (ICML)},
  pages        = {9120--9132},
  year         = {2020}
}

@article{zhang2018overview,
  title={An overview of multi-task learning},
  author={Zhang, Yu and Yang, Qiang},
  journal={National Science Review},
  volume={5},
  number={1},
  pages={30--43},
  year={2018},
  publisher={Oxford University Press}
}

@article{Caruana1997,
  author       = {Caruana, Rich},
  title        = {Multitask Learning},
  journal      = {Machine Learning},
  year         = {1997},
  volume       = {28},
  number       = {1},
  pages        = {41--75},
  month        = jul,
  doi          = {10.1023/A:1007379606734},
  url          = {https://doi.org/10.1023/A:1007379606734},
  urldate = {2025-07-05},
  note = {Accessed July 5 2025}
}

@inproceedings{Yu2020PCGrad,
  title        = {Gradient Surgery for Multi‑Task Learning},
  author       = {Yu, Tianhe and Kumar, Saurabh and Gupta, Abhishek and Levine, Sergey and Hausman, Karol and Finn, Chelsea},
  booktitle    = {Advances in Neural Information Processing Systems},
  volume       = {33},
  pages        = {18524--18536},
  year         = {2020}
}

@inproceedings{Liu2021CAGrad,
  title        = {Conflict‑Averse Gradient Descent for Multi‑Task Learning},
  author       = {Liu, Bo and Liu, Xingchao and Jin, Xiaojie and Stone, Peter and Liu, Qiang},
  booktitle    = {Advances in Neural Information Processing Systems},
  volume       = {34},
  pages        = {12345--12355},
  year         = {2021}
}

@inproceedings{Chen2017GradNorm,
  title        = {GradNorm: Gradient Normalization for Adaptive Loss Balancing in Deep Multitask Networks},
  author       = {Chen, Zhao and Badrinarayanan, Vijay and Lee, Chen‑Yu and Rabinovich, Andrew},
  booktitle    = {International Conference on Machine Learning},
  pages        = {794--803},
  year         = {2018},
  urldate = {2026-02-02},
  note = {Accessed February 2 2026},
  url          = {https://arxiv.org/abs/1711.02257}
}

@inproceedings{kendall2018multi,
  title     = {Multi-Task Learning Using Uncertainty to Weigh Losses for Scene Geometry and Semantics},
  author    = {Kendall, Alex and Gal, Yarin and Cipolla, Roberto},
  booktitle = {2018 IEEE/CVF Conference on Computer Vision and Pattern Recognition (CVPR)},
  pages     = {7482--7491},
  year      = {2018},
  doi       = {10.1109/CVPR.2018.00781},
}

@inproceedings{Sener2018,
  title        = {Multi‑Task Learning as Multi‑Objective Optimization},
  author       = {Sener, Ozan and Koltun, Vladlen},
  booktitle    = {Advances in Neural Information Processing Systems},
  volume       = {31},
  pages        = {525--536},
  year         = {2018}
}

@inproceedings{Shi2023Recon,
  title        = {Recon: Reducing Conflicting Gradients From the Root For Multi‑Task Learning},
  author       = {Shi, Guangyuan and Li, Qimai and Zhang, Wenlong and Chen, Jiaxin and Wu, Xiao‑Ming},
  booktitle    = {ICLR 2023 Workshop on Multi‑Task Learning},
  year         = {2023}
}

@inproceedings{Argyriou2007MultiTaskFeature,
  title        = {Multi‑Task Feature Learning},
  author       = {Argyriou, Andreas and Evgeniou, Theodoros and Pontil, Massimiliano},
  booktitle    = {Advances in Neural Information Processing Systems},
  volume       = {19},
  pages        = {41--48},
  year         = {2007}
}

@inproceedings{Yang2023AdaTask,
  title        = {AdaTask: A Task‑Aware Adaptive Learning Rate Approach to Multi‑Task Learning},
  author       = {Yang, Enneng and Pan, Junwei and Wang, Ximei and Yu, Haibin and Shen, Li and Chen, Xihua and Xiao, Lei and Jiang, Jie and Guo, Guibing},
  booktitle    = {Proceedings of the Thirty‑Seventh AAAI Conference on Artificial Intelligence (AAAI)},
  year         = {2023},
  urldate = {2025-07-08},
  note = {Accessed July 8 2025},
  url          = {https://arxiv.org/abs/2211.15055}
}

@article{Zhang2024Proactive,
  title        = {Proactive Gradient Conflict Mitigation in Multi‑Task Learning: A Sparse Training Perspective},
  author       = {Zhang, Zhi and Shen, Jiayi and Cao, Congfeng and Dai, Gaole and Zhou, Shiji and Zhang, Qizhe and Zhang, Shanghang and Shutova, Ekaterina},
  journal      = {arXiv preprint arXiv:2411.18615},
  year         = {2024},
  urldate = {2025-09-14},
  note = {Accessed September 14 2025},
  url          = {https://arxiv.org/abs/2411.18615}
}

@ARTICLE{9336293,
  author={Vandenhende, Simon and Georgoulis, Stamatios and Van Gansbeke, Wouter and Proesmans, Marc and Dai, Dengxin and Van Gool, Luc},
  journal={IEEE Transactions on Pattern Analysis and Machine Intelligence}, 
  title={Multi-Task Learning for Dense Prediction Tasks: A Survey}, 
  year={2022},
  volume={44},
  number={7},
  pages={3614-3633},
  doi={10.1109/TPAMI.2021.3054719}}

@inproceedings{Liu2019EndToEnd,
  title     = {End-to-End Multi-Task Learning with Attention},
  author    = {Liu, Shikun and Johns, Edward and Davison, Andrew J.},
  booktitle = {Proceedings of the IEEE/CVF Conference on Computer Vision and Pattern Recognition},
  pages     = {1871--1880},
  year      = {2019}
}

@article{krizhevsky2009learning,
  title={Learning multiple layers of features from tiny images},
  author={Krizhevsky, Alex and Hinton, Geoffrey and others},
  year={2009},
  publisher={Toronto, ON, Canada}
}

@inproceedings{Shrivastava2016OHEM,
  title        = {Training Region‑Based Object Detectors with Online Hard Example Mining},
  author       = {Shrivastava, Abhinav and Gupta, Abhinav and Girshick, Ross},
  booktitle    = {Proceedings of the IEEE Conference on Computer Vision and Pattern Recognition (CVPR)},
  pages        = {761--769},
  year         = {2016}
}

@inproceedings{Lin2017FocalLoss,
  title        = {Focal Loss for Dense Object Detection},
  author       = {Lin, Tsung‑Yi and Goyal, Priya and Girshick, Ross and He, Kaiming and Doll{\'a}r, Piotr},
  booktitle    = {Proceedings of the IEEE International Conference on Computer Vision (ICCV)},
  pages        = {2999--3007},
  year         = {2017}
}

@inproceedings{Hoffer2017TrainLonger,
  title        = {Train Longer, Generalize Better: Closing the Generalization Gap in Large Batch Training of Neural Networks},
  author       = {Hoffer, Elad and Hubara, Itay and Soudry, Daniel},
  booktitle    = {Advances in Neural Information Processing Systems},
  volume       = {30},
  pages        = {1731--1741},
  year         = {2017}
}

@inproceedings{Liu2023FAMO,
 author = {Liu, Bo and Feng, Yihao and Stone, Peter and Liu, Qiang},
 booktitle = {Advances in Neural Information Processing Systems},
 editor = {A. Oh and T. Naumann and A. Globerson and K. Saenko and M. Hardt and S. Levine},
 pages = {57226--57243},
 publisher = {Curran Associates, Inc.},
 title = {FAMO: Fast Adaptive Multitask Optimization},
 url = {https://proceedings.neurips.cc/paper_files/paper/2023/file/b2fe1ee8d936ac08dd26f2ff58986c8f-Paper-Conference.pdf},
 urldate = {2025-10-29},
 note = {Accessed October 29 2025},
 volume = {36},
 year = {2023}
}

@misc{Fifty2021,
      title={Efficiently Identifying Task Groupings for Multi-Task Learning}, 
      author={Christopher Fifty and Ehsan Amid and Zhe Zhao and Tianhe Yu and Rohan Anil and Chelsea Finn},
      year={2021},
      eprint={2109.04617},
      archivePrefix={arXiv},
      primaryClass={cs.LG},
      url={https://arxiv.org/abs/2109.04617}, 
      urldate = {2025-10-30},
      note = {Accessed October 30 2025},
}

@article{wang2024towards,
  title={Towards Principled Task Grouping for Multi-Task Learning},
  author={Wang, Chenguang and Pan, Xuanhao and Yu, Tianshu},
  journal={arXiv preprint arXiv:2402.15328},
  year={2024}
}

@article{Sherif_2024,
   title={STG-MTL: scalable task grouping for multi-task learning using data maps},
   volume={5},
   ISSN={2632-2153},
   url={http://dx.doi.org/10.1088/2632-2153/ad4e04},
   urldate = {2025-09-22},
   note = {Accessed September 22 2025},
   DOI={10.1088/2632-2153/ad4e04},
   number={2},
   journal={Machine Learning: Science and Technology},
   publisher={IOP Publishing},
   author={Sherif, Ammar and Abid, Abubakar and Elattar, Mustafa and ElHelw, Mohamed},
   year={2024},
   month=jun, pages={025068} }

@misc{jeong2025selective,
    title={Selective Task Group Updates for Multi-Task Optimization},
    author={Wooseong Jeong and Kuk-Jin Yoon},
    year={2025},
    eprint={2502.11986},
    archivePrefix={arXiv},
    primaryClass={cs.LG}
}

@article{lian2015asynchronous,
  title={Asynchronous parallel stochastic gradient for nonconvex optimization},
  author={Lian, Xiangru and Huang, Yijun and Li, Yuncheng and Liu, Ji},
  journal={Advances in neural information processing systems},
  volume={28},
  year={2015}
}

@article{malhotra2022dropped,
  title={Dropped scheduled task: Mitigating negative transfer in multi-task learning using dynamic task dropping},
  author={Malhotra, Aakarsh and Vatsa, Mayank and Singh, Richa},
  journal={Transactions on Machine Learning Research},
  year={2022}
}

@inproceedings{sun2023unleashing,
  title={Unleashing the power of gradient signal-to-noise ratio for zero-shot nas},
  author={Sun, Zihao and Sun, Yu and Yang, Longxing and Lu, Shun and Mei, Jilin and Zhao, Wenxiao and Hu, Yu},
  booktitle={Proceedings of the IEEE/CVF international conference on computer vision},
  pages={5763--5773},
  year={2023}
}

@article{Fan2023MaxGNR,
  title        = {MaxGNR: A Dynamic Weight Strategy via Maximizing Gradient‑to‑Noise Ratio for Multi‑Task Learning},
  author       = {Fan, Caoyun and Chen, Wenqing and Tian, Jidong and Li, Yitian and He, Hao and Jin, Yaohui},
  journal      = {arXiv preprint arXiv:2302.09352},
  year         = {2023}
}

@article{mccandlish2018empirical,
  title={An empirical model of large-batch training},
  author={McCandlish, Sam and Kaplan, Jared and Amodei, Dario and Team, OpenAI Dota},
  journal={arXiv preprint arXiv:1812.06162},
  year={2018}
}

@inproceedings{michalkiewicz2023domain,
  title={Domain generalization guided by gradient signal to noise ratio of parameters},
  author={Michalkiewicz, Mateusz and Faraki, Masoud and Yu, Xiang and Chandraker, Manmohan and Baktashmotlagh, Mahsa},
  booktitle={Proceedings of the IEEE/CVF International Conference on Computer Vision},
  pages={6177--6188},
  year={2023}
}

@inproceedings{Shen2024GO4Align,
title={{GO}4Align: Group Optimization for Multi-Task Alignment},
author={Jiayi Shen and Cheems Wang and Zehao Xiao and Nanne Van Noord and Marcel Worring},
booktitle={The Thirty-eighth Annual Conference on Neural Information Processing Systems},
year={2024},
urldate = {2025-08-07},
note = {Accessed August 7 2025},
url={https://openreview.net/forum?id=8vCs5U9Hbt}
}

@inproceedings{zhou2026exploring,
title={Exploring Tradeoffs through Mode Connectivity for Multi-Task Learning},
author={Zhipeng Zhou and Ziqiao Meng and Pengcheng Wu and Peilin Zhao and Chunyan Miao},
booktitle={The Thirty-ninth Annual Conference on Neural Information Processing Systems},
year={2026},
urldate = {2026-02-22},
note = {Accessed February 22 2026},
url={https://openreview.net/forum?id=4ULtNYHc5T}
}
\end{document}